\documentclass[letterpaper, 10 pt, conference]{ieeeconf}
\IEEEoverridecommandlockouts
 \pdfoutput=1 
\let\labelindent\relax         
\newcommand{\newabbreviation}[2]{#1\ (\renewcommand{#1}{#2}#1)}

\newcommand{\RMSE   }{Root Mean Square Error\xspace}

\newcommand{\LfD	}{Learning from Demonstration\xspace}

\newcommand{\TPL	}{Task Parametrised\xspace}

\newcommand{\TPGMM	}{Task-Parameterised Gaussian Mixture Model\xspace}

\newcommand{\mPGMM	}{modified Parameteric Gaussian Mixture Model\xspace}
\newcommand{\TPGMR	}{Task-Parameterised Gaussian Mixture Regression\xspace}
\newcommand{\wTPGMR	}{Relevance-Weighted Task-Parameterised Gaussian Mixture Regression\xspace}

\usepackage{amssymb}
\usepackage{mathtools}

\mathchardef\mhyphen="2D   

\newcommand{\R}     {\mathbb{R}}          

\renewcommand{\nu}  {q}                   

\newcommand{\fA} {\mathbf{A}}
\newcommand{\fb} {\mathbf{b}}

\newcommand{\statept} {\boldsymbol{\xi}} 
\newcommand{\bFr} {\mathbf{b}_{m,p}}
\newcommand{\AFr} {\mathbf{A}_{m,p}}

\newcommand{\fwgt} {\alpha}  
\newcommand{\shp} {\gamma}	

\newcommand{\Cov} {\mathbf{\Sigma}}	
\newcommand*{\sref}[1]{\S\ref{s:#1}}            
\newcommand*{\fref}[1]{Figure~\ref{f:#1}}  
\newcommand*{\eref}[1]{(\ref{e:#1})}            

\usepackage[inline]{enumitem}
\setlist{nolistsep}
\newcommand{\il}[1]{\begin{enumerate*}[label=(\roman*)]#1\end{enumerate*}}

\newcommand{\ie}{\textit{i.e.,}~} %
\newcommand{\etc}{\textit{etc.}}  %
\newlist{hypotheses}{enumerate}{10}
\setlist[hypotheses]{label*=$H_{\arabic*}$:,
                     ref=$H_{\arabic*}$}

\usepackage[normalem]{ulem}                                        
\usepackage{marginnote}
\usepackage[textwidth=10ex,colorinlistoftodos]{todonotes} 

\colorlet{jb}{red}
\colorlet{mh}{red}
\colorlet{yz}{blue}
\colorlet{as}{cyan}
\colorlet{bwm}{green}

\usepackage{color}
\usepackage{textcomp}
\usepackage{todonotes}
\usepackage[percent]{overpic}

\usepackage{wrapfig}                        
\usepackage{times}
\usepackage{graphicx}
\graphicspath{{fig/}}
\usepackage{epstopdf}
\usepackage{algorithm}
\usepackage{algorithmic}
\usepackage{multirow} 
\usepackage{multicol}
\usepackage{array}
\usepackage{mdwmath}
\usepackage{mdwtab}
\usepackage{rotating}
\usepackage[list=off,font=footnotesize]{caption}
\usepackage{amssymb}
\usepackage{verbatim}
\usepackage{subcaption}
\usepackage{pdfpages}
\usepackage{transparent}
\usepackage{xspace} 
\usepackage{color,soul} 
\usepackage{url}
\usepackage{textcomp}
\usepackage{enumitem}
\usepackage{booktabs}
\usepackage{balance}
\usepackage{dirtytalk}
\usepackage{balance}

\usepackage{times}

\usepackage{multicol}
\usepackage[bookmarks=true]{hyperref}
\usepackage{amsmath}

\makeatletter\newcommand{\manuallabel}[2]{\def\@currentlabel{#2}\label{#1}}\makeatother

\title{\LARGE \bf
Improving Task-Parameterised Movement Learning Generalisation \\with Frame-Weighted Trajectory Generation}

\author{Aran Sena, Brendan Michael
	 and Matthew Howard%
\thanks{Robot Learning Lab, Department of Informatics, King's College \mbox{London}.
        {\tt\small aran.sena@kcl.ac.uk, aransena@gmail.com}}
    \thanks{This work was funded by the Agriculture and Horticulture \mbox{Development} Board (AHDB) GROWBOT project, HNS/PO 194, and the Engineering and Physical Sciences Research Council (EPSRC) SoftSkills project, EP/P010202/1.}
         }

\begin{document}
\maketitle
\thispagestyle{empty}
\pagestyle{empty}

\begin{abstract}

\LfD depends on a robot learner generalising its learned model to unseen conditions, as it is not feasible for a person to provide a demonstration set that accounts for all possible variations in non-trivial tasks. While there are many learning methods that can handle \textit{interpolation} of observed data effectively, \textit{extrapolation} from observed data offers a much greater challenge. To address this problem of generalisation, this paper proposes a modified \TPGMR method that considers the \textit{relevance of task parameters} during trajectory generation, as determined by variance in the data. The benefits of the proposed method are first explored using a simulated reaching task data set. Here it is shown that the proposed method offers far-reaching, low-error extrapolation abilities that are different in nature to existing learning methods. Data collected from novice users for a real-world manipulation task is then considered, where it is shown that the proposed method is able to effectively reduce grasping performance errors by $\mathbf{\sim30\%}$ and extrapolate to unseen grasp targets under real-world conditions. These results indicate the proposed method serves to benefit novice users by placing less reliance on the user to provide high quality demonstration data sets.

\end{abstract}

\IEEEpeerreviewmaketitle

\section{Introduction}\noindent
This paper considers robot \newabbreviation{\LfD}{LfD\xspace}, in which examples of how to perform a task are collected from a human teacher such that the robot can use them to learn a model to perform the demonstrated task. In particular, it considers a learned model's ability to perform under situations that were not demonstrated, \ie the learners ability to \textit{generalise}, and presents a new method that significantly improves task performance in unseen conditions.

A strength often mentioned when introducing \LfD is that it enables \textit{novice users}, people who do not have the relevant knowledge to effectively program a robot, to deploy robots in labour intensive tasks by reducing the need for technical expertise \cite{Argall2009,Calinon2007b}. A corresponding weakness is then the inability for any person interacting with the robot to provide demonstrations for all conceivable variations of a non-trivial task. Furthermore, the person teaching is fallible and prone to poor teaching behaviours such as not being able to gauge the appropriate number of demonstrations required for a robot to learn a  task and struggling to identify gaps in the learners knowledge \cite{Argall2009,Sena2018TeachingLearners}. To overcome the limitations of the teacher and adapt to new situations, the robot learner must be able to effectively generalise from the demonstrations provided. 

Generalisation can take two forms, namely \il{\item \textit{interpolation}, and \item \textit{extrapolation}}. In the former, the learner must perform the task under conditions that are \textit{within} some range of conditions they have previously observed. In the latter, they must perform the task under conditions that are out-of-range of their observed experience. Many learning methods will perform well under interpolation conditions, but then degrade in performance under extrapolation \cite{Calinon2015}. Improving a robot learner's ability to extrapolate would help them to effectively learn tasks from limited demonstrations, and reduce teaching effort for their human users.

One way to improve the extrapolation ability of a learner is to consider the \textit{local structure} present in a task. For example, in learning to pick up a coffee mug, it would be beneficial for the robot to learn that the approach direction of its gripper is important for successfully grasping the mug handle. 

This approach of exploiting the local structure is used in a class of methods known as \newabbreviation{\TPL}{TP\xspace} learning, as presented in \cite{Calinon2015}. 

In \TPL learning, specific task relevant parameters are defined, such as object positions and orientations in an environment, and these are used to construct frames of reference. Data collected from the robot-point-of-view can then be ``observed'' from different points of view through these alternative frames of reference. Considering the coffee mug example, from the robot's perspective the teacher demonstrates how to reach for cups located in different locations, while from the cup's perspective, the teacher is demonstrating how to approach the cup from many start locations. By learning task representations in these local frames of reference, the learner's extrapolation abilities can be improved a great deal.

This paper presents a modified \newabbreviation{\TPGMR}{TP-GMR\xspace} method that considers the \textit{relevance of particular task parameters} before combining them into the global model. By doing so, local structure can be more effectively preserved, resulting in improved task performance. The benefits of the proposed method are shown through two experiments, highlighting the difficulty existing methods have in maintaining the local structure of a demonstrated task under extrapolation conditions. Significant improvement using the proposed method is shown in a test data set for a reaching task, versus the original \TPGMR method described in \cite{Calinon2015} and a modified \TPL method described in \cite{Calinon2013OnModels}, specifically designed to improve extrapolation of learned skills. Significant improvement is then also shown for a real-world manipulation task, with a $\sim30\%$ reduction in task error in a data set of $108$ teaching interactions with novice users.

\section{Related Work}\noindent\label{s:related}
Generalisation of demonstrated trajectories is identified as a central problem in \LfD in \cite{Osa2018AnLearningb}.
In an attempt to improve extrapolation abilities of \TPL, \cite{Calinon2013OnModels} propose a \newabbreviation{\mPGMM}{mPGMM\xspace} that uses an altered Expectation Maximisation (EM) procedure to improve the extrapolation abilities of the learnt model.
The authors show the proposed method improves the extrapolation abilities of \TPL models while reducing the computation time required compared to \mbox{\TPGMR}. While the proposed method helps in extrapolation of the demonstrated task, this method can fail to preserve local structure, resulting in failure to execute the desired task effectively.

Modification of the regression procedure is considered in \cite{Huang2018GeneralizedLearning}, where the authors propose \textit{confidence-weighted} task parameterised movement learning, but not in the context of \textit{improving} extrapolation. The authors show how different trajectories can be generated from the same set of task parameters through weighting of the learned local models. The authors propose this is a useful mechanism for providing human prior knowledge about the importance of different task parameters to the model. While this is a possible use for the method, it seems non-trivial for a person to determine the relative importance of a task frame and manually assign weightings, particularly if that person is a novice user of the technology.

In \cite{Alizadeh2014LearningParameters}, the authors discuss identifying \textit{important} task parameters for the purpose of determining whether a defined task parameter is required to learn a task or not. Here, the magnitude of the covariance matrix, evaluated using a matrix determinant, is considered an indicator for this. Under \TPGMM-based methods, variability in the recorded data is encoded in the local models' covariance matrices. Tighter groupings of data in the local models will result in ``smaller'' covariance matrices, hence the matrix determinant.

To address these issues, this paper builds on the confidence weighting scheme presented in \cite{Huang2018GeneralizedLearning} and the frame importance sampling in \cite{Alizadeh2014LearningParameters}, and presents an \textit{autonomous} method for weighting task frames during regression to improve model extrapolation, while retaining local structure observed in demonstration data.

\section{Background}\label{s:background}\noindent
Central to understanding how the proposed method improves extrapolation is understanding how local structure is modelled and used in \TPL learning methods. To this end, a brief review of the related methods is presented here.

\subsection{Task Parameterised Learning}\noindent
The learning process begins with the user providing a demonstration set consisting of $M$ demonstrations, each containing $T_m$ data points, collected in a global frame of reference. This data is formed into a data set of $N$ state measurements, $\statept_n$, with $N=\sum_{m}^{M}T_m$.

In addition to the raw data, Task Parameterised Learning builds local representations of the demonstrated task through a set of $P$ task parameters. In their most general form, these task parameters are represented by sets of affine transformations, but in the context of this work they can be simply considered as coordinate frames of reference,
\begin{equation}
p_{n,j} = \{\fb_{n}^{(j)}, \fA_{n}^{(j)}\},
\end{equation}
with $\fb$ representing the location of the frame and $\fA$ representing the orientation of the coordinate frame.

 The collected demonstrations are then projected into the local coordinate frames,
\begin{equation}
\mathbf{X}_n^{(j)} = (\fA_{n}^{(j)})^{-1}(\statept_n-\fb_{n}^{(j)}), \quad \mathbf{X}_n^{(j)} \in \R^{M\times N},
\end{equation}
where $\mathbf{X}_n^{(j)}$ is the trajectory sample at time step $n$ in frame $j$. Mixture models are then fitted to these local trajectory representations to build models of the local structure present in the data.

\subsection{Task-Parameterised Gaussian Mixture Models}\noindent
Under \TPGMM, a $K$-component mixture model is fit to the data in \textit{each} frame of reference. Each GMM consists of mixing coefficients, $\pi$, means, $\mu$, and covariances, $\Cov$. Together, these form a set of local mixture models, representing the task from multiple points of view, $\{\pi_i, \{\mu_i^{(j)}, \Cov_i^{(j)}\}_{j=1}^{P}\}_{i=1}^{K}$.

To use the local models for trajectory generation they must first be projected back into the global frame of reference and then combined into one global model. This is achieved through a linear transformation of the local models with their respective task parameters, followed by a product of Gaussians,
\begin{equation}
\mathcal{N}(\hat{\statept}_{n,i},\hat{\Cov}_{n,i}) \propto \prod\limits_{j=1}^P \mathcal{N}(\fA_{n}^{(j)}\mu_{i}^{(j)} + \fb_{n}^{(j)}, \fA_{n}^{(j)}\Cov_{i}^{(j)} \fA_{n}^{(j)})
\label{e:prod_gauss}
\end{equation}
With \eref{prod_gauss}, if the same task parameters are used in the product as were used to learn the local models, the resulting mixture model will produce a trajectory that attempts to replicate a demonstration. 

Generalisation with \TPGMM also emerges from \eref{prod_gauss}. That is, given new values for $\{\fA_j,\fb_j\}$, the local models can be used to generate global models for \textit{different} task parameters. Effectively, the local models are placed in a new pose in task space when \eref{prod_gauss} employs new task parameters. Considering task parameters as representing end-points of a trajectory, or locations of objects, it is this linear transformation followed by a product of Gaussians that allows the local models to extrapolate to new situations.

Finally, Gaussian Mixture Regression (GMR) can be used to generate a smooth path from the global $K$-component model. Assuming the model is time-driven, the GMM will encode the joint probability of states and time, $p(t,\mathbf{\statept_t})$, from which states can be sampled through GMR that computes the conditional distribution $p(\hat{\statept_t}|t)$. See \cite{Calinon2015} for more details.

\subsection{Task Parameter Weightings}\noindent
A key step in the proposed method is modifying the contribution of local models to the combined global model. A suitable method for this is proposed in \cite{Huang2018GeneralizedLearning}, with the previously mentioned confidence weighted autonomy scheme. This involves scaling the covariance matrices of a local mixture model using a weighting parameter $\fwgt$,

\begin{equation}
\mathcal{N}(\hat{\statept}_{n,i},\hat{\Cov}_{n,i}) \propto \prod\limits_{j=1}^P \mathcal{N}(\fA_{n}^{(j)}\mu_i^{(j)} + \fb_{n}^{(j)}, \fA_{n}^{(j)}\Cov_i^{(j)}/\fwgt_{n}^{(j)} \fA_{n}^{(j)})
\label{e:confidence}
\end{equation}
where $\fwgt_{n,j}$ is the weight value frame $j$ at time step $n$, with the properties $\fwgt_{n,j} \in (0,1)$. As discussed in \sref{related}, it is proposed that these weight values can be used as a method for incorporating human prior knowledge to the model; however this may be challenging for novice users. Instead, a new weighting scheme based on the relevance of a local model is proposed.

\subsection{Frame Importance}\noindent
As discussed in \sref{related}, the authors in \cite{Alizadeh2014LearningParameters} discuss identifying \textit{important} task parameters. They define the importance of frame $j$ at step $n$, $F_{n,j}$, as the ratio of the precision matrix determinant for a given frame with respect to the other frames,
\begin{equation}
F_{n,j} = \frac{|\Cov_{n,j}^{-1}|}{\sum_{j=1}^{P}|\Cov_{n,j}^{-1}|}, 
\label{e:frame_importance_old}
\end{equation}
$$
F_{n,j} \in (0,1), \quad \sum_{j=1}^{P} F_{n,j} = 1  \hspace{0.4em} \forall n.
$$

This frame importance measure will form a first step in defining the frame relevance weightings of the proposed method.
\section{Method}\label{s:method}\noindent
This section details the method used to determine optimal task parameter weightings for trajectory generation and improving the extrapolation ability of learned models. Optimisation of task parameter weightings is then shown through a task-independent, variance weighted cost function.

\subsection{Frame Relevance}\noindent
By incorporating a weighting that captures frame importance at each sample along a trajectory, the goal is to allow the global model to generate trajectory points that only consider contributions from local models \textit{when they are relevant} to the task, with the objective of improving task performance.

While the frame importance measure in \cite{Alizadeh2014LearningParameters} offers a possible solution to frame weighted trajectory generation, there are further steps that can be taken to ensure an optimal frame relevance weighting is selected. 

First, the covariance matrices used in \eref{frame_importance_old} are sampled from the learned model at the required time step through a GMR process. This has potential to introduce unwanted bias to the weightings, as a result of the choice of model parameters such as number of Gaussian components. For the purpose of determining frame relevance as indicated by demonstrations, an alternative source of information is then to directly fit a single Gaussian at each time step to the data points in each local frame of reference
\begin{equation}
\{\statept^{(j)}_{m,n}\}_{m=1}^{M} \sim \mathcal{N}(\tilde{\mu}_{n}^{(j)},\tilde{\Cov}_{n}^{(j)}).
\end{equation}
While this presents an additional computational cost, it is only required when the demonstration set is updated.

Next, instead of taking the inverse of the covariance matrices it is possible to parameterise this power, $\shp$, such that it becomes possible to {optimise} the frame weightings for a particular task
\begin{equation}
\fwgt_{n,j} = \frac{|(\tilde{\Cov}_{n}^{(j)})^\shp|}{\sum_{j=1}^{P}|(\tilde{\Cov}_{n}^{(j)})^\shp|}.
\label{e:frame_importance_new}
\end{equation}
By selecting this parameterisation, the degree to which local correlations take precedence over global correlations is controllable. As $\shp$ scales, the covariance matrices will adjust as well, such that as $\shp$ is increased trajectory generation will tend to favour the local model structure of the ``dominant'' task parameter at a given time step. For example, in a grasping task as the robot approaches the object to grab it, the model will prioritise the model in the object's local frame of reference. This is an important modification, as in a standard \TPGMM the contribution of the start position frame in \eref{prod_gauss} will offset the trajectory slightly.

A final step taken to ensure smooth transition between local models during trajectory generation is applying a smoothing procedure to the resulting frame weightings using a moving average window.

\begin{figure*}
	\vspace{0.5cm}
	\begin{scriptsize}
	\def\svgwidth{\linewidth}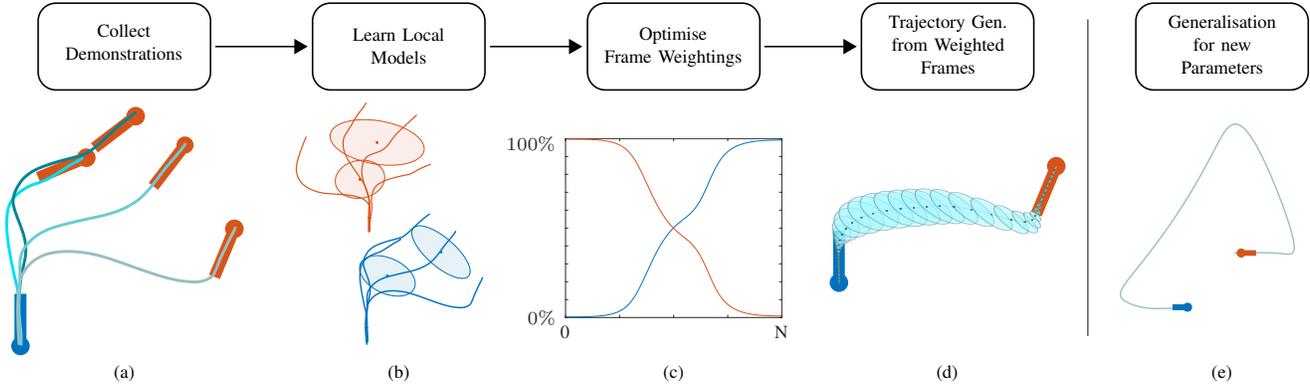
	\end{scriptsize}
	\vspace{-0.3cm}
	\caption{Data processing and model learning pipeline of the $\fwgt$TP-GMR.}%
	\label{f:pipeline}%
\end{figure*}
 
\subsection{Optimising Frame Relevance Weights}\label{s:optimisation}\noindent
Selection of $\shp$ can be optimised by treating the demonstration data set as a validation data set. Generating a set of trajectories using the task parameter sets from each demonstration will provide a set of trajectories that can be used to evaluate the learner's performance on the demonstrated task.

A \textit{weighted quadratic cost function} is defined, where the weights used directly model the variability in the data. This is done to prioritise parameter optimisation for regions of the trajectory where accuracy is required, as indicated by demonstration data variance. This is achieved by setting the weights to a diagonal matrix with entries equal to the norm of the generated data point's covariance $\Cov$, normalised such that it is in the range $(0,1)$, 
\begin{equation}
\ell = \sum_{m=1}^{M}\sum_{n=1}^{N}(\statept_{d,m,n}-\statept_{g,m,n})^\top\mathbf{W}_{m,n}(\statept_{d,m,n}-\statept_{g,m,n}),
\label{e:loss}
\end{equation}
\begin{equation}
\mathbf{W}_{m,n} = 
\begin{bmatrix}
\sigma_{m,n} & & \\
& \ddots & \\
& & \sigma_{m,n}
\end{bmatrix}, 
\end{equation} 
\begin{equation}
\sigma_{m,n} = \frac{\|\Cov_{m,n}\|}{\sum_{n=1}^{N}{\|\Cov_{m,n}\|}}. 
\end{equation} 
By defining the cost in this manner, the robot is able to prioritise its optimisation of $\shp$, with higher costs being accumulated in regions of low variability (\ie high accuracy is required), and lower costs in regions of high variability.

The optimal value for $\shp$ can then be found through a one-dimensional parameter search method. This is achieved with a bounded golden section search method, as provided in \textit{Matlab}.

The proposed method is summarised in \fref{pipeline}. This approach, with variance-adjusted frame weighting for trajectory optimisation, forms the \newabbreviation{\wTPGMR}{$\fwgt$TP-GMR\xspace} method.

\section{Experiments}\label{s:exp}\noindent
The proposed approach is first evaluated on a test data set for a reaching task, followed by performance evaluation on a real-world manipulation task with a more complex state representation. The objective in these experiments is to explore the proposed method's ability to extrapolate tasks to unseen conditions, and show how this ability is of practical use in a real-world scenario.

\subsection{Reaching Task Performance}\label{s:exp1}\noindent
This first experiment investigates the performance of \wTPGMR on a reaching task data set, compared to two contemporary methods, \TPGMR and \mPGMM. 

The reaching task data set used in this experiment\footnote{Reaching data set available from \url{http://www.idiap.ch/software/pbdlib/}\cite{Calinon2015}}, consists of four demonstrations showing a point-to-point reaching task that approximates removing an end-effector from one pocket and inserting it into another (shown in \fref{pipeline}(a)). There are two sets of task parameters. The first parameter for each demonstration forms a coordinate frame centred on its start location, with the orientation aligned with the direction of travel. The second forms a coordinate frame centred on the goal location with a fixed orientation (red and blue markers in \fref{pipeline} respectively). 

\subsubsection{Setup}\noindent
 In the data set, the state consists of a time index, and the location of the trajectory point, \mbox{$\statept_n = (t_n, x_n, y_n)^\top$},  where for the state at each sample $n$, $t$ is the time step. The task frames, $\{\fb,\fA\}$, are then defined as follows,
\begin{equation}
\bFr = \begin{pmatrix}0,  x_{m,p},  y_{m,p}  \end{pmatrix}^{\top}, \quad
\AFr = \begin{pmatrix}1 & \mathbf{0} \\ \mathbf{0} & \mathbf{R}_{m,p} \\ \end{pmatrix}, 
\end{equation}
where $(x_m,y_m)$ is the position of the $p^{th}$ frame for the $m^{th}$ demonstration, and $\mathbf{R}_{m,p} \in \R^{2\times2}$ is a planar rotation matrix representing the orientation of the frame. The task frames are static over time steps, but vary per demonstration $m$.  

In addition to task parameters, $K=3$ components are used for each of the models. For \wTPGMR, confidence weightings are estimated for the frames following the procedure described in \sref{method}.

Evaluation of a model's ability to learn the task performance is achieved through an exhaustive \textit{leave-one-out} cross-validation procedure. For each model option, \TPGMR, \mPGMM, and \wTPGMR, a model is learned using $M-1$ of the available demonstrations. The reproduction score for the selected model is then taken as the \newabbreviation{\RMSE}{RMSE\xspace} between the set-aside trajectory and a trajectory generated from the learnt model with the remaining set-aside trajectory's task parameters. This procedure is repeated, cycling which demonstration is left out and resetting the model on each attempt, until each demonstration has been used as the cross-validation test trajectory.

\subsubsection{Results \& Discussion}\noindent
Table \ref{table:exp1results} shows the results from this initial test with the point-to-point reaching data set. 

It can be seen that of the three methods tested, \wTPGMR incurs the lowest error for this data set. This indicates that \wTPGMR was the most accurate in generating trajectories for unseen conditions, albeit over a small range.

\begin{table}[ht]
	\centering
	\scriptsize
	\caption{Exhaustive leave-1-out cross-validation results for \sref{exp1}.}
	\begin{tabular}[t]{cccc}
		\toprule
		~	& \TPGMR  & \wTPGMR   & \mPGMM \\ \midrule
		RMSE & 0.279 &  0.197 & 0.270 \\ 
		Std. & $\pm$0.146 & $\pm$0.112 & $\pm$0.140 \\
		\bottomrule
	\end{tabular}
	\label{table:exp1results}
\end{table}

\begin{figure}
	\vspace{0.2cm}
	\begin{scriptsize}
		\def\svgwidth{\linewidth}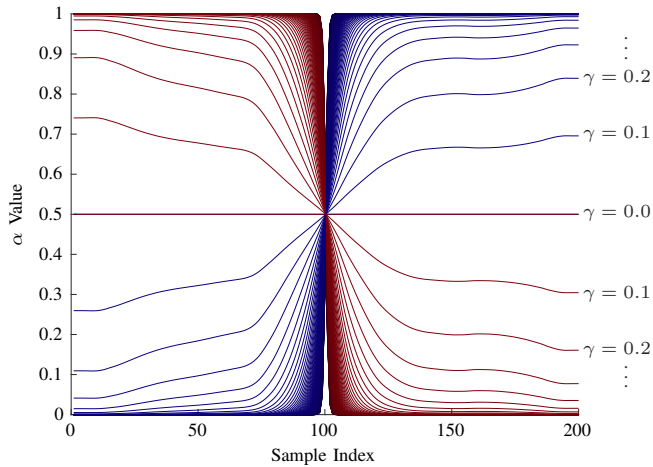
	\end{scriptsize}
	
	\caption{Effect of  $\shp$ on $\fwgt$ for the two-frame model used in \sref{exp1}. Lines indicate frame relevance over the course of a trajectory from the demonstration test data set, with red representing Frame 1, and blue Frame 2.}%
	\label{f:alpha}%
\end{figure}

Looking more closely at the learning process for \wTPGMR, \fref{alpha} shows plots of frame weightings generated for the reaching task model, with increasing values of $\shp$. Here, the red lines indicate the start frame, and the blue lines indicate the goal frame. As a frame weighting value increases, it decreases the corresponding covariance as defined in \eref{confidence}, as expected. Initially, the first frame takes priority followed by a transition to the second frame as the trajectory approaches the goal. The key observation here is that, for increasing $\shp$, the frame weighting will increasingly favour one local model over the other. At $\shp=0$, each frame is given equal weighting $\fwgt_j = 0.5$; however as $\shp$ increases, the transition from one frame to another becomes increasingly steep. It is the controlability of this transition that allows the \wTPGMR method to optimise trajectory generation effectively.

 \begin{table}[ht] 
 \centering 
 \caption{Summary statistics for grid-search test case where the starting frames location is varied, but its rotation is similar to rotations observed in the demonstration set.}
 \scriptsize 
 \begin{tabular}[t]{rcccccc} 
 \toprule 
~&\multicolumn{2}{c}{\textbf{Constraint Errors}} & \multicolumn{2}{c}{\textbf{Task Errors}} & \multicolumn{2}{c}{\textbf{Path Lengths}}\\
~ & \textbf{Mean}  & \textbf{Std.} & \textbf{Mean}  & \textbf{Std.}& \textbf{Mean}  & \textbf{Std.}\\ \midrule 
\TPGMR & 19.34 & $\pm$2.59 & 1.10 &  $\pm$0.74 & 9.53 & $\pm$3.68 \\ 
\wTPGMR & 1.00 &  $\pm$0.00 & 0.04 &  $\pm$0.00 & 9.73 &  $\pm$3.40 \\ 
\mPGMM & 18.00 &  $\pm$2.76 & 1.03 &  $\pm$1.30 & 10.66 &  $\pm$4.84 \\ 
\bottomrule 
 \end{tabular} 
 \label{table:exp2results} 
 \end{table}

\begin{figure}
	\vspace{0.2cm}
		\scriptsize
	\def\svgwidth{0.95\linewidth}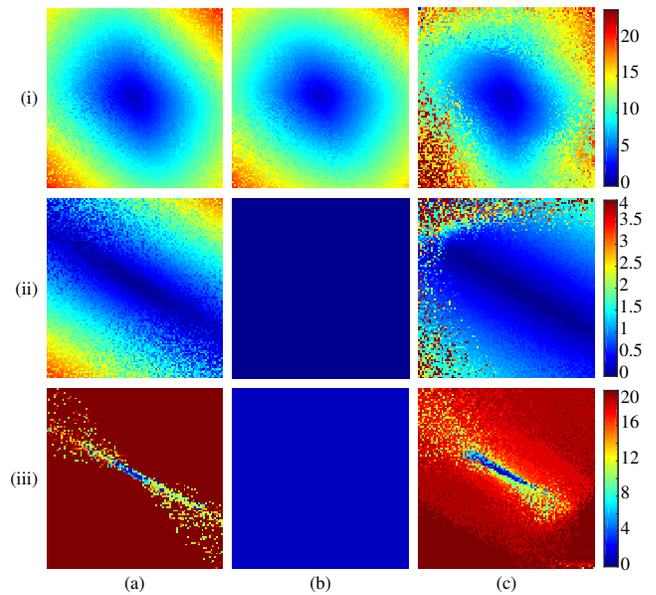
	\caption{Generalisation test results for three learning methods. Column (a) \TPGMR, (b) \wTPGMR, (c) \mPGMM. Row (i) total path length, (ii) Euclidean distance error at start and end of trajectory, (iii) Constraint satisfaction at start and end of trajectory.}  %
	\label{f:comps}%
\end{figure}

\subsection{Extrapolation Performance}\noindent
The second experiment investigates the method's ability to extrapolate task performance to unseen conditions. This is shown through a grid search approach that expands far around the original demonstration area, where trajectories are generated with the learnt model for a series of starting positions.

\subsubsection{Setup}
Here, a $10\textrm{m}\times10$m grid of task parameters is constructed, centred on $(0,0)$. Each parameter in the grid is given an orientation similar to those encountered in the demonstration set. Each of these start position parameters is then paired with a goal parameter that is the same as the one used in the demonstrations. A large degree of extrapolation is considered here. In the original data set, the goal location is set at $(-0.8, -0.8)$ and each of the start task parameters are located $\sim1.5$m away from the goal.

For each parameter set in the grid, a trajectory is generated and evaluated on three criteria. The criteria are \il{\item trajectory length, \item trajectory end-points error, \item constraint satisfaction error.} Trajectory length is taken as an indicator of the quality of the demonstration, as longer trajectories can be an indicator of incoherent paths and shorter trajectories can be an indicator of incomplete paths. Trajectory end-point error specifically evaluates the model's ability to generate a trajectory that starts and ends where it is meant to. As found in \cite{Calinon2015}, task modelling methods that do not exploit local structure can rapidly see this type of error increase, potentially resulting in erratic movement at the start of a trajectory and incomplete actions due to movements ending early. 

The constraint satisfaction error is designed to capture the model's ability to generate paths that exit and enter the start and end frames in the correct orientation. From the start frame of reference, trajectories should move in the direction that the frame is pointing until the path is clear of the frame marker. For the goal frame of reference, trajectories should enter directly down from the top of the frame marker. Pragmatically, this is evaluated by counting how many data points are within bounding boxes placed at each end of the generated trajectory. These bounding boxes are chosen such that the first and last 10 data points of each demonstration trajectory are contained within them. The error count is taken as absolute value, so that the learning method will be penalised for too many as well as too few data points being located in these bounding boxes. If 10 data points are counted in each bounding box, it is assumed that the trajectory satisfied the task constraint.

\subsubsection{Results \& Discussion}\noindent
\begin{figure}
	\vspace{0.15cm}
	\scriptsize
	\def\svgwidth{\linewidth}
\begingroup%
  \makeatletter%
  \providecommand\color[2][]{%
    \errmessage{(Inkscape) Color is used for the text in Inkscape, but the package 'color.sty' is not loaded}%
    \renewcommand\color[2][]{}%
  }%
  \providecommand\transparent[1]{%
    \errmessage{(Inkscape) Transparency is used (non-zero) for the text in Inkscape, but the package 'transparent.sty' is not loaded}%
    \renewcommand\transparent[1]{}%
  }%
  \providecommand\rotatebox[2]{#2}%
  \newcommand*\fsize{\dimexpr\f@size pt\relax}%
  \newcommand*\lineheight[1]{\fontsize{\fsize}{#1\fsize}\selectfont}%
  \ifx\svgwidth\undefined%
    \setlength{\unitlength}{583.46827229bp}%
    \ifx\svgscale\undefined%
      \relax%
    \else%
      \setlength{\unitlength}{\unitlength * \real{\svgscale}}%
    \fi%
  \else%
    \setlength{\unitlength}{\svgwidth}%
  \fi%
  \global\let\svgwidth\undefined%
  \global\let\svgscale\undefined%
  \makeatother%
  \begin{picture}(1,0.48218132)%
    \lineheight{1}%
    \setlength\tabcolsep{0pt}%
    \put(0,0){\includegraphics[width=\unitlength,page=1]{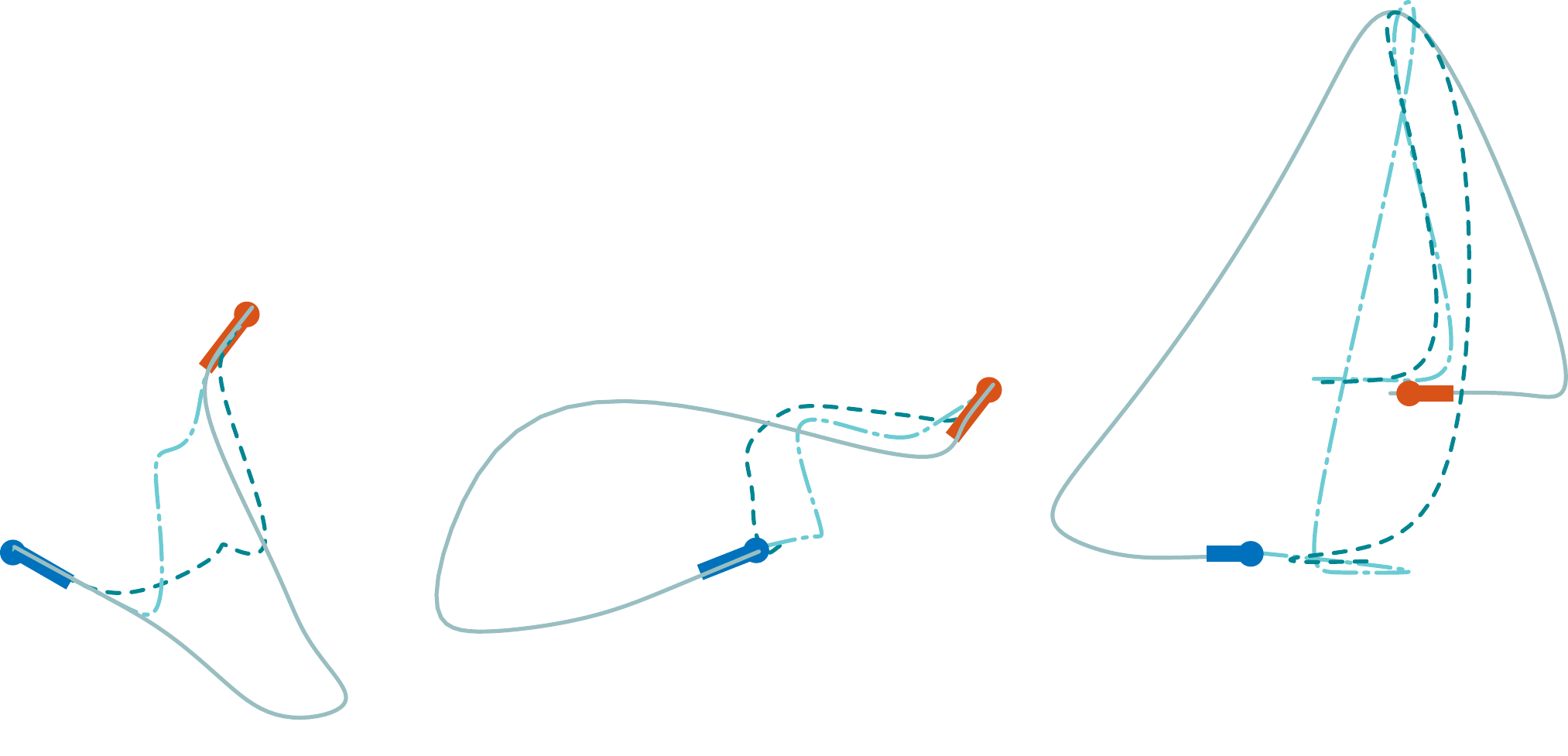}}%
    \put(0.11127616,0.00158165){\color[rgb]{0,0,0}\makebox(0,0)[t]{\lineheight{1.25}\smash{\begin{tabular}[t]{c}(a)\end{tabular}}}}%
    \put(0.45772463,0.00489139){\color[rgb]{0,0,0}\makebox(0,0)[t]{\lineheight{1.25}\smash{\begin{tabular}[t]{c}(b)\end{tabular}}}}%
    \put(0.8350224,0.00670931){\color[rgb]{0,0,0}\makebox(0,0)[t]{\lineheight{1.25}\smash{\begin{tabular}[t]{c}(c)\end{tabular}}}}%
    \put(0.10661182,0.3936459){\color[rgb]{0,0,0}\makebox(0,0)[lt]{\lineheight{1.25}\smash{\begin{tabular}[t]{l}TP-GMR\end{tabular}}}}%
    \put(0,0){\includegraphics[width=\unitlength,page=2]{generalisation.pdf}}%
    \put(0.10661182,0.35853271){\color[rgb]{0,0,0}\makebox(0,0)[lt]{\lineheight{1.25}\smash{\begin{tabular}[t]{l}mPGMM\end{tabular}}}}%
    \put(0,0){\includegraphics[width=\unitlength,page=3]{generalisation.pdf}}%
    \put(0.10661182,0.42711543){\color[rgb]{0,0,0}\makebox(0,0)[lt]{\lineheight{1.25}\smash{\begin{tabular}[t]{l}$\alpha$TP-GMR\end{tabular}}}}%
  \end{picture}%
\endgroup%

	\caption{Extrapolation tests comparing the proposed method against alternative models under three conditions. (a) Goal frame rotated $120$ degrees, (b) Goal frame rotated $240$ degrees, (c) Goal and origin frame rotated $90$ and $270$ degrees respectively. The generated trajectories must start and end at the red and blue frames respectively, with the paths exiting and entering along the direction indicated by the frame indicator. Note that only the proposed method succeeds in all cases.\vspace{-0.5cm}}%
	\label{f:gen_test}%
\end{figure}

The results in Table \ref{table:exp2results} highlight the significant difference in performance between the three methods. Here, it can be seen that \wTPGMR achieves much lower task and constraint error values. This indicates that under \wTPGMR, the model is able to generate trajectories that accurately produce the requested path, and importantly this path follows local structure constraints, as provided in the original demonstrates.

Plots of the results are provided in \fref{comps}, where each method occupies one column, and each criteria is plotted along one row. The distance criteria is plotted along row (i). It can be seen that for (a) \TPGMR and (b) \wTPGMR that the generated path lengths are largely similar; however for (b) \mPGMM, the trajectories begin to become erratic in the extremes of the grid.

Looking at row (ii), end-point error, it can be seen that the performance of (a) \TPGMR degrades in a regular pattern as trajectories move further away from the original demonstration set. It can be seen that \mPGMM provides improved performance over \TPGMR; however in the upper and lower left corners the trajectory generation becomes erratic in a manner similar to the first row. In addition to erratic end-point error, it can be seen that the error in (c) does increase with a regular pattern like (a), albeit to a lesser degree. Plot (b) presents the first unusual result, where it can be seen that the error for \wTPGMR is very low and constant across the grid. This result is made possible by the clean data in the data set producing frame weightings that accurately prioritise one frame over the other.

Looking at the final row in \fref{comps}, task constraint errors, reveals another unusual result for \wTPGMR. In (a), \TPGMR can be seen to have a very small low-error region, which directly lines-up with the original demonstration region. From this, it can be concluded that \TPGMR is only effective in the neighbourhood of the original demonstrations, given the patterns seen in end-point error and constraint error. In (c), \mPGMM does not fare much better than \TPGMR and similar conclusions can be drawn. Finally in (b), it can be seen that again there is a low, constant level of error across the grid.

The combination of low end-point error, low task-constraint error, and a smoothly increasing path length is a powerful combination. While these strong results are largely due to the clean nature of the data set, they are indicative of the ability for \wTPGMR to greatly enhance the extrapolative abilities of \LfD systems.

\fref{gen_test} presents some samples from each of the models in a variety of generalisation challenges. It can be seen that in each case, \wTPGMR is able to generate a smooth trajectory which satisfies the task constraints.

These results raise the question of whether an extrapolated trajectory is correct and should be used. If the model can produce an accurate trajectory to a previously unseen scenario, there is uncertainty over whether this is a safe or correct action to take.  Some steps to automate detection of uncertain states can be found in \cite{Maeda2017ActivePrimitives}; however whether or not to trust the system largely remains at the discretion of the user. Ultimately, if a trajectory is suitably extrapolated, and the person agrees, this presents a large time saving for them.

\begin{figure}
	\vspace{0.3cm}
\begin{center}
	\begin{overpic}[width=0.95\linewidth]{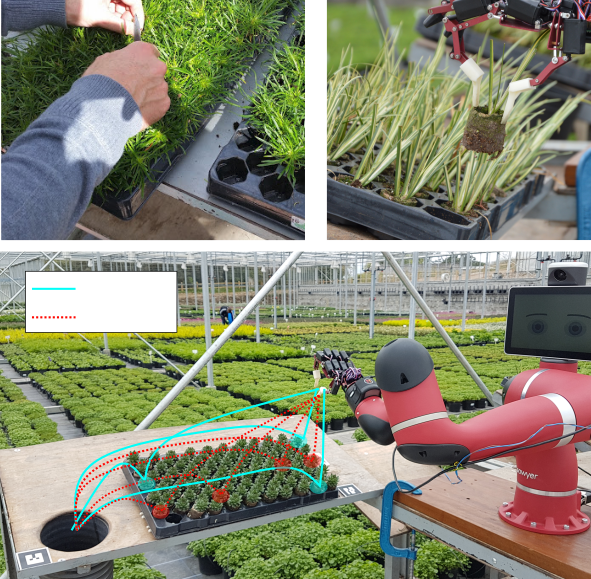}
		\put(13.4,48.1) {\scriptsize Demonstrated}
				\put(13.4,43.3) {\scriptsize Generated}	
		\end{overpic}
		\end{center}
	\caption{Horticultural sorting task used in \sref{exp-real}. Here, unhealthy plants must be removed from a plant tray. Key to this is the ability to generalise from user provided demonstrated trajectories (cyan), to determine the appropriate trajectory to pick plants from previously unseen locations (red).}%
	\label{f:gex}%
\end{figure}
\begin{figure*}
	\centering
	\vspace{.5cm}
	\scriptsize
	\def\svgwidth{0.95\linewidth}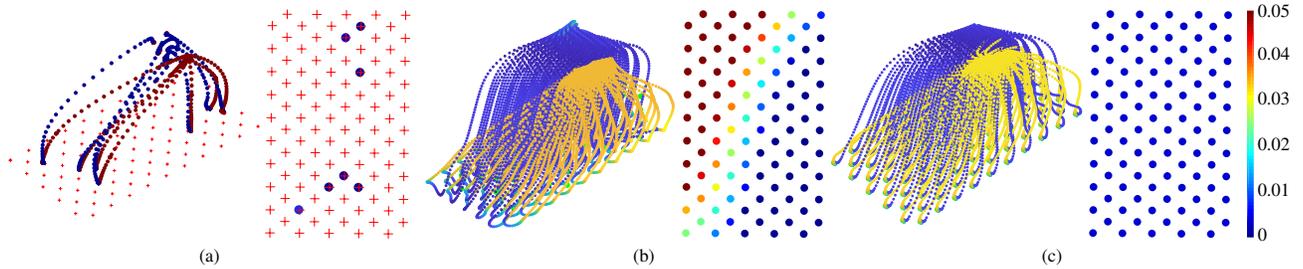
	\caption{Demonstration data (a) provided by novice users on the real world horticultural task shown in \fref{gex}. (b) Displays the trajectories generated using \TPGMR. (c) Displays the trajectories generated using \wTPGMR. The colour of the dots in the grid plots indicates the positioning error of the trajectory when executing a grab action for that dot. The color of the trajectories indicates the hand control signal state, with the displayed plots showing the gripper beginning open and closing once it is at the grasp target.}  %
	\label{f:real_data_plots}%
\end{figure*}
\subsection{Real World Manipulation Task}\label{s:exp-real}\noindent
Having seen the benefits of the proposed method on the reaching task data set, a real-world task is considered that presents further challenges.

In this experiment, the data used was collected from a real-world robot system, with demonstrations provided by \textit{novice users} (\ie people who do not have prior knowledge of robotics, machine learning, \LfD, \etc)\footnote{This experiment was conducted with ethical approval granted by KCL REC Committee under LRS-17/18-5549.}\footnote{A DOI to this data set will be made available in the final paper.}. The robot used is a Rethink Robotics Sawyer, with an Active8 AR10 hand, and the task under consideration is a horticultural sorting task as found in mass production sites of ornamental plants, where rejected products must be removed from a tray and discarded, see \fref{gex}.

\LfD is useful for this task, as there can be a great deal of variety in the production process on grower sites. There can be hundreds of varieties of plants grown at one site over the course of a year, with different plants requiring different manipulation strategies. Additionally, the plants and flowers are grown in a variety of pots and tray ranging in capacity from 25-100 plants. In this scenario, a learning method that is able to \textit{accurately} generalise from a few demonstrations to many locations would be a great help in providing flexible automation.

\subsubsection{Setup}\label{s:exp2-setup}\noindent
In this experiment, \TPGMR and \wTPGMR are used to learn models of the task. Specifically, the task involves learning to pick up, remove, and place a plant from a tray of $100$ to a disposal bucket. The objective provided to the participants was to teach the robot to perform the disposal task for any plant position in the tray. This task was demonstrated by $36$ participants, $3$ times each, providing $108$ teaching interactions to consider. 

The state and task parameters are then defined as follows,
\begin{align}
\statept_n &= \begin{bmatrix}t_n \\ \mathbf{x}_n^p \\ \mathbf{x}_n^q \\ x_n^h \end{bmatrix} \in \R^{8\times1}, \\
\bFr &= \begin{bmatrix}0 \\ \mathbf{p}_{m,j} \\ \mathbf{0} \end{bmatrix}  \in \R^{8\times1},
\AFr = \mathbf{I}^{8\times8} 
\end{align}
where $\mathbf{x}_n^p$ and $\mathbf{p}_{m,j}$ are the positions of data point $n$ and frame $j$ for demonstration $m$, respectively, and $\mathbf{x}_n^q$ is the orientation of data point $n$ demonstration $m$ using an axis-angle representation. $x_n^h$ is a scalar control signal used to open and close the robot hand, and $\mathbf{I}^{8\times8}$ is the identity matrix.

The performance of the robot learner under the two learning methods is then evaluated by generating a test set of trajectories for each of the $100$ plant positions in the tray. Each trajectory is evaluated by comparing the end-effector position at critical points during the task execution. 

These critical points are \il{\item the start location, \item the grab location, and \item the place location.} Position (i) ensures that the model is generating a trajectory that starts where it is meant to, thus avoiding sudden jerks in movement at the start of the task. Position (ii) is evaluated by identifying the location of the robot hand at the point it closes its gripper, and comparing it to the mean location of the hand during grasping in the demonstration set. Position (iii) then ensures that the robot is correctly depositing the picked plant, and is identified as the point at which the robot opens its hand. 

In evaluating the learning methods in this way, it is assumed that the demonstration data provides correct information on how to pick the plants from the tray, and that if the local structure of the generated trajectory (\ie the grabbing location relative to the plant) does not closely match the demonstration data, then the robot will fail to grasp the plant.

\subsubsection{Results \& Discussion}\noindent
Analysing the collected data revealed the distribution of data residuals was found to be non-normal by an Anderson-Darling test on the data. Considering this non-normality, the data was tested using a paired non-parametric Wilcoxon Signed-Rank test. This indicated that the median error \wTPGMR ($\tilde{\epsilon} = 5.0541$) was statistically significantly lower than under conventional \TPGMR ($\tilde{\epsilon} = 3.4763$), $Z = 8.6621, p < 10^{-17}$.

This result confirms the findings observed in the previous two experiments. Note that in all experiments, while the learning methods were assessed using task-specific criteria, the learning process used the task-independent cost function \eref{loss} described in \sref{optimisation}.

Looking more closely at the results and plotting a selection of the data generated by \TPGMR and \wTPGMR reveals some useful details. A demonstration set is shown in  \fref{real_data_plots}(a) along with its corresponding grasp-error plot. When using \TPGMR to learn from this data set, as shown in \fref{real_data_plots}(b), this data set can be seen to be suboptimal. There are two sub-groups of demonstrations in the top and bottom portion of the tray, which results in redundant demonstrations in the demonstrated regions, and undemonstrated states elsewhere. By switching to a \wTPGMR learning mode, with no adjustment to the demonstration set, \fref{real_data_plots}(c) shows a large improvement in the performance of the learner. In particular, it can be seen that in \fref{real_data_plots}, the trajectories near the grasp targets closely match the trajectories in the demonstration set, indicating that the local structure has been learned and is being used.

This is an important result in \LfD. Given that people often struggle to provide adequate demonstration sets \cite{Argall2009,Sena2018TeachingLearners}, a learning method that can effectively extrapolate from regions that have been shown could reduce the challenge of providing good demonstration sets for \LfD.

Note that unlike in the first set of reaching task experiments, the error achieved does not reduce to a constant level. This is due to the noisy nature of the data making learning more challenging, and resulting in less information being available to the robot learner to gauge which frames are important at each step.

\section{Conclusion} \noindent
\label{sec:conclusion}
Extrapolation in \LfD presents many challenges and opportunities. As discussed in related work, and shown through experiments in \sref{exp1}, prior approaches to improving extrapolation in Task Parameterised learning have limited generalisation abilities beyond the original demonstrations. This paper presents a new approach, Relevance-Weighted Task-Parameterised Gaussian Mixture Regression, for addressing this problem. Under this method, task parameters are modulated based on their estimated importance during each time step in a trajectory.

As demonstrated in a series of experiments with both simulated data for a reaching task, and real world data collected from novice users, this approach significantly improves the extrapolation abilities of \TPGMR. These improvements will serve to benefit novice users of \LfD systems, by enhancing the ability of robot learners to extrapolate from limited data and places less reliance on the user providing a high-quality demonstration data sets.

Limitations of the proposed approach include the issue of time distortion. Generated trajectories have the same number of data points as original demonstrations, so additional processing may be required to ensure robot limits are not exceeded. There is also the more fundamental question of how to determine whether an extrapolated trajectory is correct and should be trusted. Whilst this is a common consideration when extrapolating using any learning method, the high degree of extrapolation possible with \wTPGMR might mislead a novice user to be confident in a generated trajectory that is unsuitable.

Future work could consider the effect of \wTPGMR on users presented with extrapolated trajectories. Further, the discussed models represent a small subset of models available in \TPL. It would be of interest to explore how relevance based frame weighting can be applied to state-based systems, and tasks with force interactions.

\section*{Acknowledgments}\noindent
The authors wish to thank the many people who took time to participate in this experiment.

\bibliographystyle{IEEEtran}
\bibliography{references}

\end{document}